\documentclass[journal]{IEEEtran}

\usepackage{graphicx}
\usepackage{float}
\usepackage{color}

\usepackage{amssymb}
\usepackage{algorithm}
\usepackage{algorithmic}
\usepackage{times}  
\usepackage{helvet}  
\usepackage{courier}  
\usepackage{url}  
\usepackage{graphicx}  
\usepackage{epsfig}

\ifCLASSOPTIONcompsoc
\usepackage[caption=false,font=normalsize,labelfont=sf,textfont=sf]{subfig}
\else
\usepackage[caption=false,font=footnotesize]{subfig}
\fi

\begin{document}

\title{A Novel Brain Decoding Method: a Correlation Network Framework for Revealing Brain Connections}

\author{Siyu Yu, Nanning Zheng$^\star$, \emph{Fellow, IEEE}, YongQiang Ma, Hao Wu, and Badong Chen, \emph{Senior Member, IEEE}
\thanks{S. Yu, N. Zheng($^\star$correspondence author) Y. Ma, H. Wu and B. Chen are with the Institute of Artificial Intelligence and Robotics, Xi'an Jiaotong University, Xi'an, Shaanxi 710049, P.R. China. E-mail: {yusiyu, musaqiang, xuan.zhi}@stu.xjtu.edu.cn, {nnzheng, chenbd}@mail.xjtu.edu.cn.}}


\maketitle

\begin{abstract}
Brain decoding is a hot spot in cognitive science, which focuses on reconstructing perceptual images from brain activities. Analyzing the correlations of collected data from human brain activities and representing activity patterns are two problems in brain decoding based on functional magnetic resonance imaging (fMRI) signals. However, existing correlation analysis methods mainly focus on the strength information of voxel, which reveals functional connectivity in the cerebral cortex. They tend to neglect the structural information that implies the intracortical or intrinsic connections; that is, structural connectivity. Hence, the effective connectivity inferred by these methods is relatively unilateral. Therefore, we proposed a correlation network (CorrNet) framework that could be flexibly combined with diverse pattern representation models. In the CorrNet framework, the topological correlation was introduced to reveal structural information.  Rich correlations were obtained, which contributed to specifying the underlying effective connectivity. We also combined the CorrNet framework with a linear support vector machine (SVM) and a dynamic evolving spike neuron network (SNN) for pattern representation separately, thus providing a novel method for decoding cognitive activity patterns. Experimental results verified the reliability and robustness of our CorrNet framework and demonstrated that the new method achieved significant improvement in brain decoding over comparable methods.
\end{abstract}


\IEEEpeerreviewmaketitle

\section{Introduction}

\IEEEPARstart{O}{nly} with  a good  understanding of the brain can we develop more robust artificial intelligence (AI) \cite{zheng2017hybrid}. Regrettably, the cognitive mechanism of the human brain remains unclear. Brain decoding that focuses on reconstructing perceptual images from brain activities is faced with great challenges. In recent decades, brain decoding with functional magnetic resonance imaging (fMRI) signals has driven plenty of studies \cite{haxby2001distributed,kamitani2005decoding}.
%
%
Most relevant methods are based on multi-voxel pattern analysis \cite{hausfeld2014multiclass} or voxel-wise modeling \cite{naselaris2015voxel}.
%
%
%
The main idea is the decoding of fMRI signals evoked by perceptual stimuli to obtain brain activity patterns, and representing these patterns for perceptual image reconstruction \cite{norman2006beyond,kay2008identifying}.
Hence, there are two primary problems in brain decoding with severe noise and high-dimensional fMRI signals. One is how to thoroughly analyze the correlations of collected data, and the other is how to effectively represent the brain's activity patterns, which can rely on powerful classifiers in machine learning \cite{horikawa2017generic}. We believe that the former is more critical to cognitive science.

\begin{figure*}[htbp]
  \begin{center}
  \includegraphics[width=0.95\textwidth]{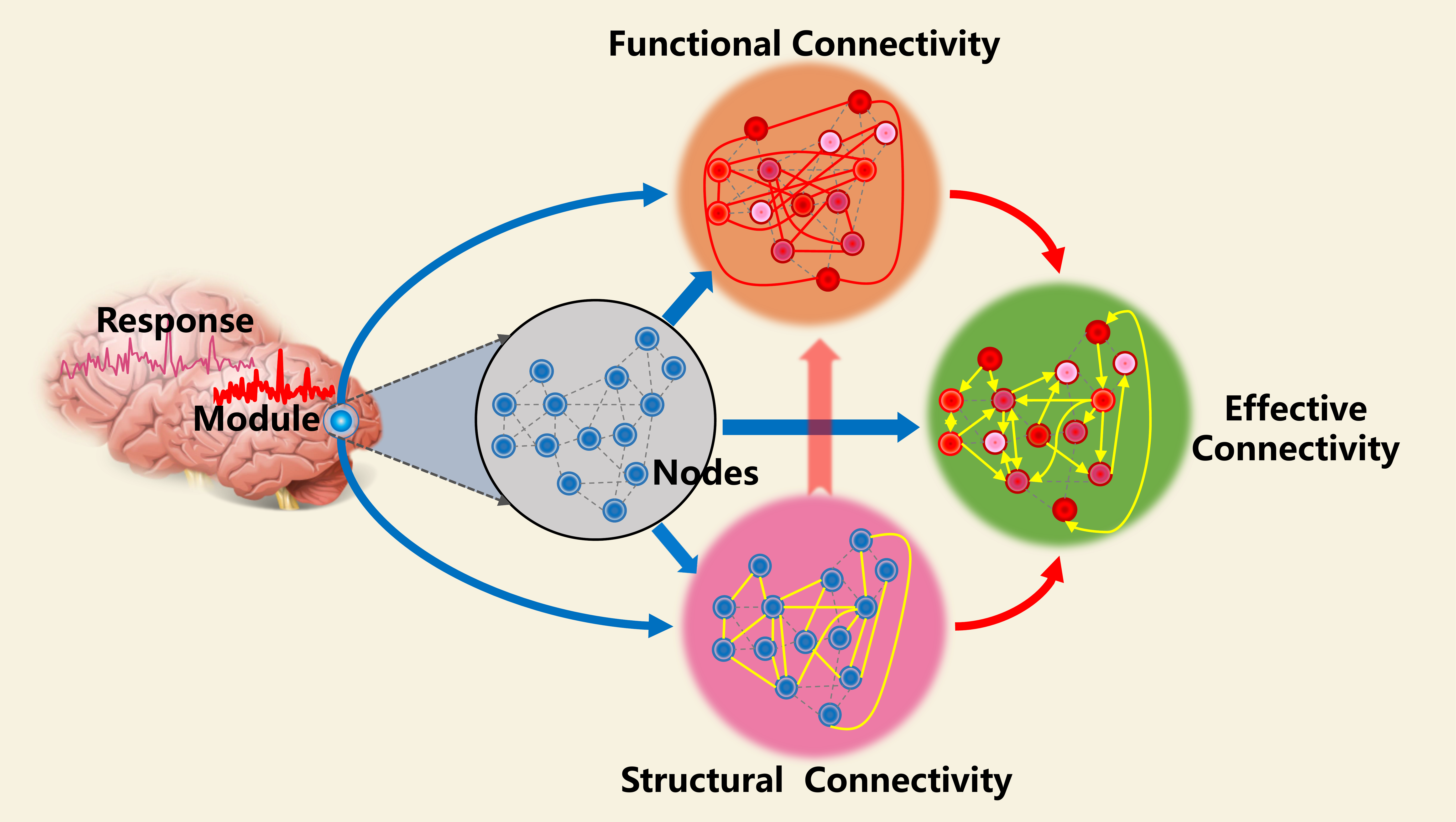}\\
  \caption{The three types of connections between brain nodes: functional connectivity, structural connectivity, and effective connectivity. The big blue point in the brain presents a certain module that contains some nodes (small points), and we highlight the module in the gray circle.  The dotted gray lines between nodes in the four circles are the imaginary connections used to shape the module. In terms of functional connectivity, nodes in different colors are of different response values. The more red the node, the stronger is its response. Thick red lines are used to connect nodes with the same response strength, revealing functional connectivity. In the pink circle, the thick yellow lines connect nodes that are adjacent according to anatomy. In the green circle, effective connectivity can be inferred by yellow one-sided or two-sided arrows. The one-sided arrows represent the structure-function relationships between two adjacent nodes with different response patterns. The arrow direction means a strong-response node can affect its neighborhood. When two adjacent nodes have the same response pattern, there is a potential interaction effect; this is indicated by the two-sided arrows.}\label{fig:connect}
  \end{center}
\end{figure*}

Complex brain activity states with limited data instances make it necessary to analyze the correlations between fMRI signals and visual stimuli. As is well known, the human brain encodes perceptual images into hierarchical signals, while the reconstruction task aims to decode the brain's responses to the stimuli; hence, the reconstruction process reverts to the process of the human brain's mechanism.   
In terms of studies on structure-function relationships from the view of brain network theory, it is believed that, as Fig. \ref{fig:connect} shows, there are three types of connections between brain nodes: functional connectivity, structural connectivity, and effective connectivity \cite{park2013structural}. Actually, the brain nodes are the basic elements of brain network studies, such as neurons in anatomy or voxels in fMRI experiments.
%
%
%
%
The correlations in reconstruction tasks are expected to correspond to the three kinds of connectivity, even for simulating and modeling them, so that the correlation analysis can contribute to revealing the working mechanism of the human brain.
An earlier study proposed a modular decoding approach \cite{miyawaki2009visual}. Later, Bayesian canonical correlation analysis (BCCA) \cite{fujiwara2013modular} was proposed for brain decoding.
Meanwhile, generative models with variational autoencoders (VAEs) \cite{kingma2013auto} were introduced, such as deep canonically correlated autoencoders (DCCAE) \cite{wang2015deep}.
Nevertheless, most existing correlation models \cite{woolrich2004fully,kuang2014discrimination} may be poor in capturing the correlations because they tend to focus mainly on the strength information of voxel, which reveals the functional connectivity in the cerebral cortex, while neglecting the structural information of voxel, implying the intracortical or intrinsic connections, i.e., the structural connectivity.
An early study \cite{zeeman1965topology} proved that adjacent neurons in the same cortex tend to present similar activity patterns; that is to say, the functional similarity between neurons is dependent on the structural connectivity. Notably, the effective connectivity expected to be inferred in experiments is exactly constrained by structural connectivity. Therefore, the effective connectivity specified by previous reconstruction methods seems not to be comprehensive and have limited performance. Thus, it is necessary to analyze both the strength values of voxel and the structural information to enrich the correlations and explore the effective connectivity in attention tasks (perceptual tasks). This is the main contribution of this paper.

Moreover, suitable classifiers are required to reveal brain activity patterns as pattern representation models on the basis of the correlations, overcoming the perplexing measurement noise. It is hard to make a reliable choice from the numerous machine learning algorithms. Existing studies generally use simple methods (linear models) because fMRI signals have severe noise, are high-dimensional, and often involve small sample sizes. However, most linear reconstruction models may be weak in feature extractions of brain activity patterns. Although some complicated pattern representation models, like DCCAE, have been used and have achieved relatively effective performance, they require many intermediate variables and strong hypotheses, thus bringing severe computational problems. Obviously, a single deep learning (DL) model, which has been a topic of active research in recent years, is not suitable for analyzing brain signals with a small sample size in the early visual cortex. A single DL model can result in over-fitting.
%
%
There are still no powerful, general methods for perceptual image reconstruction for both correlation analysis and brain activity pattern representation.
Support vector machines (SVMs) \cite{laconte2005support} have been used to efficiently represent activity patterns; they are powerful for pattern recognition and classification. Therefore, SVM was one of our optional pattern representation methods.
Recently, researchers have tried to use more biological models, such as spike neuron networks (SNNs) \cite{maass1996lower}, to reconstruct perceptual images \cite{kasabov2014neucube,kasabov2017mapping}. Compared with artificial neural networks (ANNs), the learned connections of SNNs demonstrate the dynamic spatiotemporal relationships in fMRI signals; however, there have not been many studies on this to date. Since we believe that SNNs can be promising for brain decoding, this was the other choice for the pattern representation process in the novel method.

In this paper, we proposed a novel brain decoding method composed of a self-adaptive correlation network (CorrNet) framework and a pattern representation model.
The CorrNet framework was established by considering three correlations (the correlation among brain activity, the correlation between brain activity and visual stimuli, and the correlation among visual stimuli), and could be combined with diverse pattern representation models. In the CorrNet framework, firstly, the topological correlation and strength correlation were considered.  
%
%
Then, a probabilistic correlation graph in a joint space was generated by iteratively updating the CorrNet framework in a self-learning manner to obtain the pixel-wise correlation pairs (pixel–voxel pairs). Next, the pattern representation model was trained to learn connection weights from the correlation pairs to represent brain activity patterns.
Specifically, the linear SVM and the dynamic evolving SNN were both used as pattern representation models separately. This could verify the reliability and robustness of our CorrNet framework.
Finally, we tested our method and other methods (Miyawaki\cite{miyawaki2009visual} and BCCA \cite{fujiwara2013modular}) on a new fMRI dataset from the early visual cortex (mainly area V1).
Experimental results demonstrated that our method achieved better performance than the other methods in brain decoding, and the CorrNet framework was reliable and robust.

The rest of this paper is organized as follows: Section $2$ reviews some related works about brain decoding. Then, the proposed method is introduced in Section $3$. Section $4$ reports the experimental results. Finally, our conclusion is given in Section $5$ as well as some directions for future work.
\section{Related Works}
Much of the literature in cognitive science is devoted to brain decoding.
An earlier method combined multi-scale local image bases with multiple scales to decode brain activities \cite{miyawaki2009visual}.
Furthermore, Bayesian inference algorithms and correlation neural networks have been combined \cite{fujiwara2013modular,hossein2016reconstruction} to analyze the correlations. The existing correlation methods were derived from a voxel receptive-field model and only relied on the strength information to locate a small perceptual stimulus that elicits activity spatially spread over voxels \cite{engel1997retinotopic,shmuel2007spatio}.
%
%
As for pattern representation, machine learning algorithms are used, such as linear regression \cite{yamashita2008sparse,yamashita2009sparse}, Bayesian classifiers \cite{flandin2007bayesian,penny2011statistical}, and linear observation models \cite{beckmann2003general,schoenmakers2013linear}. These methods generally have limited feature extraction power. In addition, some researchers attempted to use deep VAEs, like DCCAE. However, models with VAEs require many intermediate variables and strong hypotheses, leading to poor generalization and high computational cost.
Some researchers found that the linear SVM combined with multivariate voxel selection is effective in classifying fMRI spatial patterns \cite{de2008combining}. Recently, a spatiotemporal data machine of evolving SNNs was proposed to represent activity patterns of fMRI signals \cite{kasabov2017mapping}.

Notably, theories about structural and functional brain networks offer insight into connection-to-cognition modeling \cite{park2013structural,sporns2013network}.
%
%
Inspired by the cognitive studies, in this paper we proposed a CorrNet framework that could be combined with diverse pattern representation models.
The CorrNet framework reserves the advantages of past correlation analysis methods in learning the relationships between the strength of fMRI signals and visual stimuli, which reveal the functional connectivity in the human brain. Meanwhile, it considers structure information of voxels by introducing the topological correlation, thus specifying the structural connectivity. Therefore, the effective connectivity inferred by the CorrNet framework to resolve the dialectic between structure and function is more reliable and comprehensive than previous models.
For pattern representation, both a linear SVM model and an SNN model were separately combined with the CorrNet framework, proposing a new method to reconstruct visual images. Our method could avoid many intermediate variables and unnecessary strong hypotheses. To the best of our knowledge, this paper is the first to study topological correlation in perceptual image reconstruction and to discuss the three types of connections (functional connectivity, structural connectivity, and effective connectivity) in attention tasks.

\section{Proposed Method}

Our work rests on how the early visual cortex organizes local interactions to deal with different perceptual stimuli on the basis of function–structure relationships in the brain. In other words, the proposed method was expected to reveal the functional connectivity by the correlation between brain activities based on blood oxygenation level-dependent (BOLD) fMRI signals acquired during perceptual task performance. Additionally, we attempted to specify the structural connectivity by using the adjacent relationships of voxels in the cortex, instead of analyzing diffusion MRI signals or anatomy.
Therefore, the effective connectivity could be inferred by using our model-based method to estimate the model parameters (weight values) that can best reconstruct the perceptual images. As long as we can model the brain activity patterns from the obtained effective connectivity, human-like computer vision in the complex visual scenario may no longer be a mystery.
%
%
¡¾¼ÓÍ¼¡¿
%
%
Actually, for attention tasks, like perceptual tasks, existing studies have not found a reliable way to analyze the structure correlation. Besides, in past reconstruction experiments that only considered the function correlation, the effective connectivity between voxels was not comprehensive due to the lack of structure correlation; this led to unsatisfactory experimental results. In this paper, we tried to compensate for these past studies by introducing a topological model into our CorrNet framework. All brain decoding methods can use our proposed CorreNet framework in the correlation analysis stage. Therefore, to some extent, our work explains and simulates the brain's cognitive mechanism. 
%
%

\begin{figure}[h]
  \begin{center}
  \includegraphics[width=0.47\textwidth]{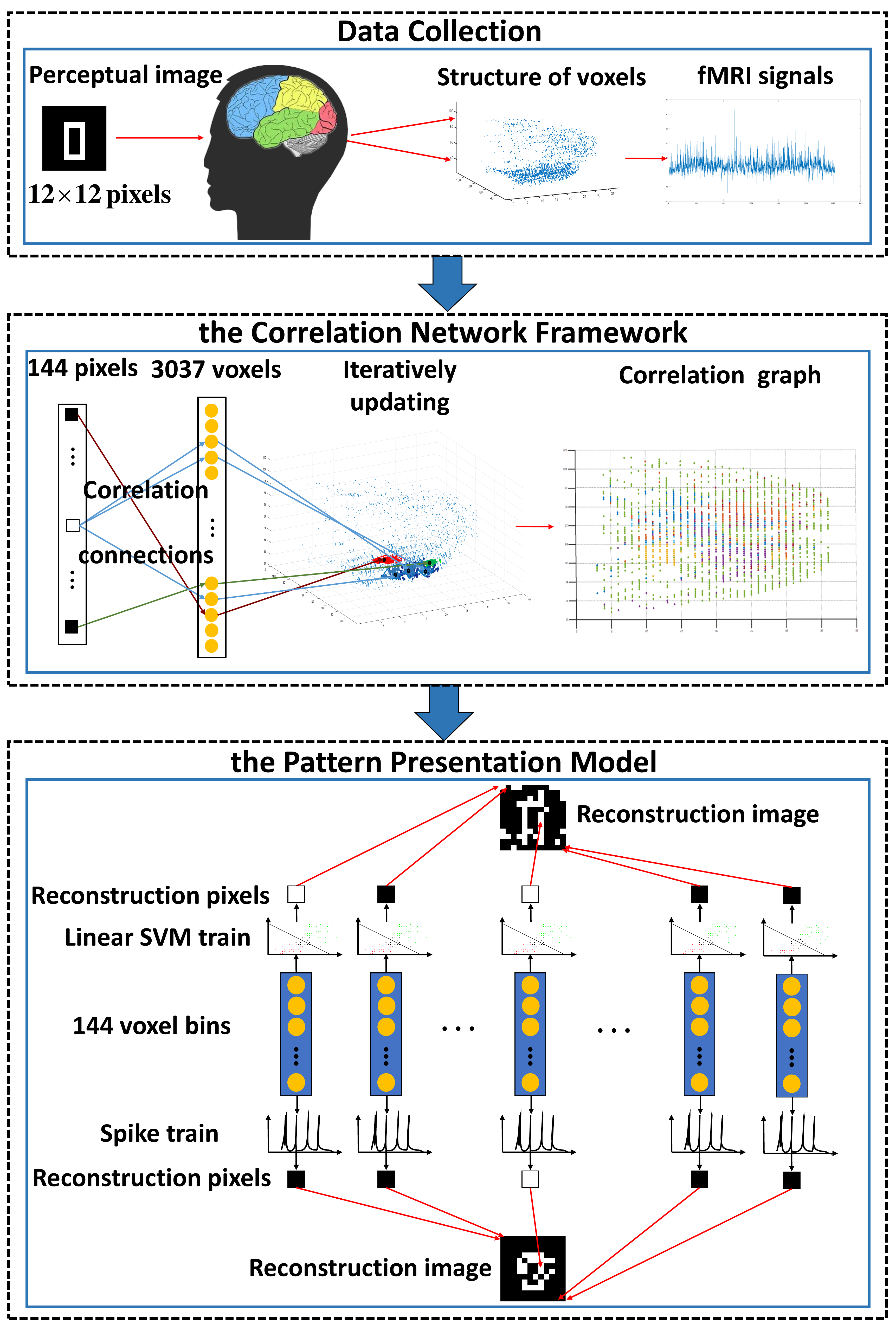}\\
  \caption{Flow chart of the proposed method. The complete process of brain decoding contains three steps. In data collection, stimuli, fMRI signals, and the brain structure information of subjects are collected. Then, the CorrNet framework analyzes the data correlations. Finally, with suitable classifier models, such as SVM and SNN, the activity patterns can be represented.}\label{fig:liuc}
  \end{center}
\end{figure}

As Fig. \ref{fig:liuc} shows, the proposed method contained two parts in terms of the two primary problems in brain decoding. One is the CorrNet framework for obtaining the correlations of fMRI signals and perceptual images; the other is the pattern representation model for learning brain activity patterns by training a classifier on the basis of the CorrNet framework.
The procedure is simple:
First, there are two distinct views $(\bf{X,Y})$, denoted by $({{\bf{x}}_1},{{\bf{y}}_1})$,..., $({{\bf{x}}_N},{{\bf{y}}_N})$, where $N$ is the number of training trials, ${{\bf{x}}_i}\in{\mathbb{R}^{{D_1}}}$, and ${{\bf{y}}_i}\in{\mathbb{R}^{{D_2}}}$ for $i = 1,...,N$.
${\bf{X}}\in{\mathbb{R}^{{D_1} \times N}}$ and ${\bf{Y}}\in{\mathbb{R}^{{D_2} \times N}}$ denote fMRI signals with ${D_1}$ voxels and visual images with ${D_2}$  pixels. Besides, ${{\bf{pos}}_j} \in {\mathbb{R}^{{3}}}$ for $j = 1,...,{D_1}$ denotes the position of the $j^{th}$ voxel in the three-dimensional world coordinate frame.
In addition, ${{\bf{v}}_j} \in {\mathbb{R}^N},j = 1,...,{D_1}$ and ${{\bf{p}}_k} \in {\mathbb{R}^N},k = 1,...,{D_2}$ denote the strength value vector of the $j^{th}$ voxel and the pixel value vector of the $k^{th}$ pixel, respectively.
Based on the pixel-wise inverse receptive field, the CorrNet framework was generated to obtain the correlation voxel bins, denoted as ${\bf{bi}}{{\bf{n}}_k}$ for $k = 1,...,{D_2}$.
${\bf{bi}}{{\bf{n}}_k}{\rm{ = }}\{ {{\bf{v}}_j}{\rm{|}}{\bf{Corr}}({{\bf{v}}_j},{{\bf{p}}_k}){\rm{ = 1}},j \in \{ 1,...,{D_1}\} \}$
presents the set of voxels that are in the inverse receptive field of pixel ${{\bf{p}}_k}$, where ${\bf{Corr}}({{\bf{v}}_j}{\rm{ , }}{{\bf{p}}_k}{\rm{)}}$ is the correlation matrix.
${\bf{Corr}}({{\bf{v}}_j}{\rm{ , }}{{\bf{p}}_k}{\rm{) = 1}}$ indicates that ${\bf{v}}_j$ is in the inverse receptive field of ${\bf{p}}_k$; otherwise, ${\bf{Corr}}({{\bf{v}}_j}{\rm{ , }}{{\bf{p}}_k}{\rm{) = 0}}$.
Then, based on pixel-wise modeling, we used a distributed pattern representation model to build the maps from ${\bf{bi}}{{\bf{n}}_k}$ to ${\bf{p}}_k$ directly:
\begin{equation}\label{equ:ECS}
{{\bf{p}}_k} = {f^{(k)}}({\bf{bi}}{{\bf{n}}_k}),k = 1,...,{D_2},
\end{equation}
where $f^{(k)}$ denotes the reconstruction function of ${\bf{p}}_k$. Since each pixel is reconstructed by the pixel-wise representation method, the reconstruction image is available.

\subsection{Correlation Network Framework}

\begin{figure}[htbp]
  \begin{center}
  \includegraphics[width=0.48\textwidth]{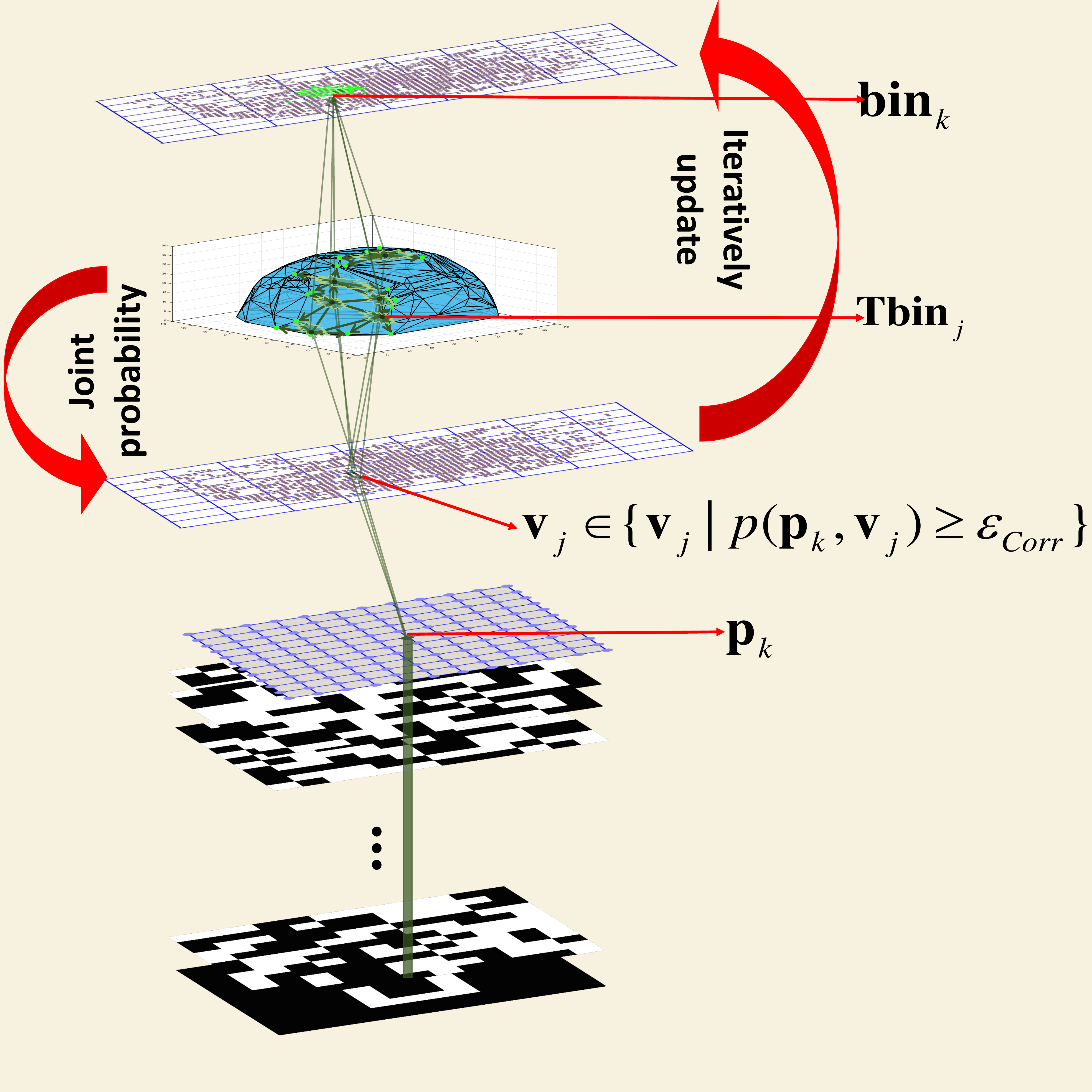}\\
  \caption{The updating procedure of the CorrNet framework includes the following: generating the topological structure to obtain the topological neighborhood ${\bf{Tb}}{{\bf{in}}_j}$ of ${\bf{v}}_j$; finding the strength correlation set of ${\bf{p}}_k$; using the topological correlation and strength correlation to update the probabilistic correlation graph; and getting the ${\bf{b}}{{\bf{in}}_k}$.}\label{fig:cor}
  \end{center}
\end{figure}
The CorrNet framework is one of the contributions in this paper. As Fig. \ref{fig:cor} shows, the CorrNet framework was proposed to generate the correlation bins ${\bf{bi}}{{\bf{n}}_k}$ and the correlation matrix ${\bf{Corr}} \in {\mathbb{R}^{{D_1} \times {D_2}}}$ from the stimuli–response pairs.
There are three kinds of correlations considered in this paper: the correlation among brain activities, the correlation between brain activities and visual stimuli, and the correlation among visual stimuli. We believe the three kinds of correlations correspond to the three kinds of connections mentioned above; that is, functional connectivity, structural connectivity, and effective connectivity.
In the CorrNet framework, when analyzing the strength information of voxels under diverse stimuli, the correlation between brain activities and visual stimuli can be obtained to reveal the functional connectivity in the early visual cortex.
Once we modeled the topological correlation underlying the structure information of voxels, the structural connectivity could be viewed. Therefore, the effective connectivity could be inferred by the CorrNet framework, which adequately considers the correlations of collected data and contains the reliable model parameters helping to best reconstruct the perceptual images.

First, the correlation among visual stimuli were visualized as reconstruction results and verified through experiments.
Then, we generated a probabilistic graph in a joint space to analyze the the correlation among brain activities as well as the correlation between brain activities and visual stimuli. In detail, the correlation among brain activities is denoted as a binary matrix ${\bf{Cv}} \in {\mathbb{R}^{{D_1} \times {D_1}}}$. ${\bf{Cv}}$ is determined by the two factors of the voxels (strength information $S$ and structure information $POS$, which is measured in the three-dimensional world coordinate frame):
\begin{equation}
\begin{array}{ll}
{\bf{Cv}}(j,m) &= \left\{ \begin{array}{ll}
1&{,~p(S,POS)_{j,m} \ge {\varepsilon _{Cv}}}\\
\\
0&{,~\rm{otherwise}}
\end{array} \right.\\
\\
p(S,POS) &= p(S|POS)p(POS),
\end{array}
\end{equation}
where $p(POS)\in {\mathbb{R}^{{D_1} \times {D_1}}}$ represents the topological correlation and is viewed as the prior probability. The computation procedure will be detailed subsequently. ${\varepsilon _{Cv}}\in [0,1]$ is an adjustable parameter and was set to 0.1 in this paper. $p(S|POS)\in {\mathbb{R}^{{D_1} \times {D_1}}}$ denotes the strength correlation measured by a full covariance matrix:
\begin{equation}
p{(S|POS)_{j,m}}{\rm{ = }}\frac{{E({{\bf{v}}_j}{{\bf{v}}_m}) - E({{\bf{v}}_j})E({{\bf{v}}_m})}}{{\sqrt {D({{\bf{v}}_j})} \sqrt {D({{\bf{v}}_m})} }},
\end{equation}
where $D({{\bf{v}}_j})$ is the variance of ${\bf{v}}_j$.

\subsubsection{Topological correlation}

Because the function of certain brain regions are inevitably dependent on structure, the diverse responses of neurons rely on their various distributions in the cortex. An earlier study proved that adjacent neurons tend to have similar receptive fields and activity patterns \cite{zeeman1965topology,sporns2013network}.
%
%
However, almost all existing reconstruction methods have neglected the structure information of voxels.
Hence, to enrich the correlations in brain decoding, we analyzed the structure information of voxels and obtained the prior probability $p(POS)$.
Because there are complex drapes, hierarchies, and numerous neurons in the cortex, it seems impossible to unfold the visual cortex to model the spatial relationship neuron-by-neuron. In this paper, spatial variation was introduced by utilizing the three-dimensional position in the world coordinate frame. In fact, there is exactly a bijective mapping between the voxel position in the cortex and that in the three-dimensional world coordinate frame.
Because the visual cortex is a two-dimensional overlapped structure, two voxels that are adjacent in the three-dimensional world coordinate frame may be non-neighboring in the cortex. Inversely, it can be ascertained that voxels adjacent in the cortex are bound to be neighboring in the three-dimensional world coordinate frame.
In summary, because the spatial variation from the cortex to the world coordinate frame may shorten the distance between voxels, the topological neighborhood of a certain voxel in the cortex is the subset of that in the world coordinate frame. Based on this conclusion, we computed $p(POS)$ by using the position ${\bf{pos}_j}$ for $j = 1,...,{D_1}$ directly in the world coordinate frame.

In the world coordinate frame, an undirected graph $G = \{ \mathbb{V},E\}$ was generated by Delaunay triangulations to present the topological (structural) connections of all voxels \cite{chew1989constrained}.
$\mathbb{V}$ is the set of all voxels ${\bf{v}}_j$, and $E$ presents the set of valid Delaunay edges, which define the topological correlations among voxels.
Notably, the reason why we use Delaunay triangulations for the voxels instead of simply the physical distance within the brain is because Delaunay triangulations can reveal hierarchical (superficial) connectivity, consistent with the hierarchical structure of the cerebral cortex. Therefore, the topological method, instead of the anatomy method, was utilized to simulate the structural connectivity in the brain. Only based on the hierarchical connectivity revealed by Delaunay triangulations can the physical distance be meaningful to the simulation. These Delaunay triangulations guarantee that the nearest neighbor graph is a subgraph of the Delaunay triangulation.
Thus, the topological neighborhood of the $j^{th}$ voxel can be found and is denoted as ${\bf{Tb}}{{\bf{in}}_j}$ for $j = 1,...,{D_1}$. ${\bf{Tb}}{{\bf{in}}_j}$ is defined as
${\bf{Tbi}}{{\bf{n}}_j}{\rm{ = }}\{ {{\bf{v}}_m}{\rm{|}}E({\bf{po}}{{\bf{s}}_j},{\bf{po}}{{\bf{s}}_m}){\rm{ = 1,}}m \in \{ 1,...,{D_1}\} {\rm{\} }} \cup \{ {{\bf{v}}_l}{\rm{| }}d(j,l) \le {\varepsilon _{{d_j}}},l \in \{ 1,...,{D_1}\} \}, $
where $d(j,l)$ presents the Euclidean distance between ${\bf{v}}_j$ and ${\bf{v}}_l$ in the world coordinate frame, and $\varepsilon _{{d_j}}$ is a self-adaption distance parameter and varies from diverse pattern representation models. The value of $\varepsilon _{{d_j}}$ was used to adjust the influence of the topological correlation in the CorrNet framework, and a large value means that the topological correlation has a high influence. Finally, $p(POS)$ was obtained as below.
\begin{equation}
p{(POS)_{j,m}} = \left\{ \begin{array}{ll}
1 &{,{{\bf{v}}_m} \in {\bf{Tbi}}{{\bf{n}}_j}}\\
\\
0 &{,\rm{otherwise}}
\end{array} \right.
\end{equation}

\subsubsection{Probabilistic correlation graph}

The objective correlation matrix $\bf{Corr}$  was determined by the correlation pairs $({{\bf{p}}_k},{\bf{bi}}{{\bf{n}}_k})$ for $k = 1,...,{D_2}$ in a  probabilistic correlation graph, denoted ${G_{Corr}} = \{ \mathbb{VP},{E_{Corr}}\} $. $\mathbb{VP}$ is the set of all voxels and pixels. $E_{Corr}$ represents the weighted edges that define the correlation connections between ${\bf{v}}_j$ and ${\bf{p}}_k $. The weight values are equal to the probability that ${\bf{v}}_j$ belongs to ${\bf{bi}}{{\bf{n}}_k}$, denoted as $p({{\bf{p}}_k},{{\bf{v}}_j}|{\bf{bi}}{{\bf{n}}_k})$, which can be expressed by Bayes rules:

\begin{equation}
p({{\bf{p}}_k},{{\bf{v}}_j}|{\bf{bi}}{{\bf{n}}_k}){\rm{ =}}\frac{{p({\bf{bi}}{{\bf{n}}_k}|{{\bf{p}}_k},{{\bf{v}}_j})p({{\bf{p}}_k},{{\bf{v}}_j})}}{{p({\bf{bi}}{{\bf{n}}_k})}},
\end{equation}
where $p({\bf{bi}}{{\bf{n}}_k}|{{\bf{p}}_k},{{\bf{v}}_j})$ is dependent on the topological correlation and will be detailed later; $p({\bf{bi}}{{\bf{n}}_k}){\rm{ = }}1$ is fixed and means there must exist ${\bf{bi}}{{\bf{n}}_k}$ for ${\bf{p}}_k$.
$p({{\bf{p}}_k},{{\bf{v}}_j})$ is defined by a full covariance matrix:
\begin{equation}
p({{\bf{p}}_k},{{\bf{v}}_j}){\rm{ = }}\frac{{\sum\limits_{i = 1}^N {({v_{i,j}} - \bar v)({p_{i,k}} - \bar p)} }}{{N - 1}}.
\end{equation}
Then, in the CorrNet framework, the weights $p({{\bf{p}}_k},{{\bf{v}}_j}|{\bf{bi}}{{\bf{n}}_k})$ are iteratively updated. The procedure is shown in Algorithm.\ref{alg:level}.
After the iteration, if $p({{\bf{p}}_k},{{\bf{v}}_j}|{\bf{bi}}{{\bf{n}}_k}) \ge {\varepsilon _{Corr}}$, ${\bf{v}}_j$ is in the inverse receptive field of ${\bf{p}}_k$; namely, ${\bf{Corr}}({{\bf{v}}_j}{\rm{ , }}{{\bf{p}}_k}{\rm{) = 1}}$ , and ${\bf{v}}_j\in {\bf{bi}}{{\bf{n}}_k}$. ${\varepsilon _{Corr}}\in [0,1]$ is a correlation parameter and was set to 0.5 in this paper.

\begin{algorithm}[t]
\caption{ Generating the Probabilistic Graph $G_{Corr}$.}
\label{alg:level}
\begin{algorithmic}[1] 
\REQUIRE ~~\\ 
Matrix ${\bf{Cv}}$, the correlation parameter $\varepsilon _{Corr}$, and the strength correlation $p({{\bf{p}}_k},{{\bf{v}}_j})$ 
\\
\ENSURE ~~\\ 
The correlation probability $p({{\bf{p}}_k},{{\bf{v}}_j}|{\bf{bi}}{{\bf{n}}_k})$ 
\STATE Initial $p({\bf{bi}}{{\bf{n}}_k}|{{\bf{p}}_k},{{\bf{v}}_j})=1$ and $p({{\bf{p}}_k},{{\bf{v}}_j}|{\bf{bi}}{{\bf{n}}_k})=p({{\bf{p}}_k},{{\bf{v}}_j})$ and $L^{(k)}=\emptyset$ ($k=1\rightarrow {D_2},j=1\rightarrow {D_1}$)
\FOR{$k=1$ to ${D_2}$}
\FOR{$j=1$ to ${D_1}$}
\IF{$p({{\bf{p}}_k},{{\bf{v}}_j}) \ge {\varepsilon _{Corr}}$}
\STATE
Update ${L^{(k)}} = {L^{(k)}} \cup \{ m|{\bf{Cv}}(j,m) = 1\}$;\\
${k_t} = \mathop {\min }\limits_{m \in {L^{(k)}},m \ne j} d({{\bf{v}}_j},{{\bf{v}}_m})$;\\
for all $m\in {L^{(k)}}$, calculate: 
\\ ~~~~$p({\bf{bi}}{{\bf{n}}_k}|{{\bf{p}}_k},{{\bf{v}}_m})={k_{t}}/d({{\bf{v}}_j},{{\bf{v}}_m})$;\\
~~~~${p_{t}} = p({{\bf{p}}_k},{{\bf{v}}_m})\times p({\bf{bi}}{{\bf{n}}_k}|{{\bf{p}}_k},{{\bf{v}}_m})$;\\
~~~~$p({{\bf{p}}_k},{{\bf{v}}_m}|{\bf{bi}}{{\bf{n}}_k}){\rm{ = }}\max (p({{\bf{p}}_k},{{\bf{v}}_j}|{\bf{bi}}{{\bf{n}}_k}),{p_{t}})$;
\ENDIF
\ENDFOR
\IF{$L^{(k)}==\emptyset$}
\STATE
$j=\mathop {\arg \max }\limits_{{j^ * } \in \{ 1,...,{D_1}\} } p({{\bf{p}}_k},{{\bf{v}}_{{j^ * }}}|{\bf{bi}}{{\bf{n}}_k})$;\\
Return to step 5;
\ENDIF
\STATE ${\bf{bi}}{{\bf{n}}_k}={L^{(k)}}$;\\
\ENDFOR
\end{algorithmic}
\end{algorithm}

\subsection{Pattern representation model}

Most reconstruction methods tend to model on the stimulus–response pair $(\bf{X,Y})$ and decode the most probable image $\bf{Y}$ from the fMRI signal $\bf{X}$. In this paper, the pixel-wise retinotopy in V1 is utilized to build a map from the voxel bin ${\bf{bi}}{{\bf{n}}_k}$ to its correlation pixel ${\bf{p}}_k$ directly, finding the corresponding activity patterns. For pattern representation, the linear SVM has been proved to be effective. In addition, compared with ANNs, SNNs are more biomimetic in both information transmission and structure.
Therefore, both the linear SVM model and SNN model were trained to learn brain activity patterns from the correlation pairs, verifying the robustness of the CorrNet framework. In the training process, our models served as distributed multi-voxel receptive-field models.
The activity patterns were learned through constructing the maps in pairs $({{\bf{p}}_k},{\bf{bi}}{{\bf{n}}_k})$. ${\bf{bi}}{{\bf{n}}_k}$ was the input, and ${\bf{p}}_k$ was the expected output.
The reconstruction procedure involved monitoring ${\bf{bi}}{{\bf{n}}_k}$ corresponding to the retinotopy map of ${\bf{p}}_k$ by inversing the receptive-field model.
Compared with conventional retinotopy, our pixel-wise method solved the problem of missing activity patterns by using the multi-voxel correlation bin ${\bf{bi}}{{\bf{n}}_k}$.
Meanwhile, the proposed CorrNet framework took the various correlations into full consideration, guaranteeing the accuracy of the reconstruction models.
The submodels in Equation \ref{equ:ECS} worked in parallel, and the pair $({{\bf{p}}_k},{\bf{bi}}{{\bf{n}}_k})$ contained $N$ trials in the training process.

\subsubsection{Linear support vector machine}

We assumed that the pattern representation model was given by the linear SVM to find a linear decision boundary in the feature space, such that
\begin{equation}
{{\bf{p}}_k} = {\bf{W}}_{\bf{k}}^{\bf{T}}{{\bf{X}}^{(k)}} + {w_0},k = 1,...,{D_2},
\end{equation}
where ${\bf{W}}_{\bf{k}}^{\bf{T}}$ defines the linear decision boundaries, and each attribution of ${{\bf{X}}^{(k)}}{\rm{ = [}}{\bf{x}}_1^{(k)}{\rm{,}}{\bf{x}}_2^{(k)}{\rm{,}}...{\rm{,}}{\bf{x}}_N^{(k)}{{\rm{]}}^T}$. ${\bf{x}}_i^{(k)}$ for $i=1,...,N$ is the strength value vector of the voxels in ${\bf{bi}}{{\bf{n}}_k}$ for the $i^{th}$ trial.
${\bf{W}}_{\bf{k}}^{\bf{T}}$ is dependent on the support vectors of ${\bf{bi}}{{\bf{n}}_k}$, which are exactly active voxels corresponding to a certain activity pattern.
Therefore, the linear SVM model could highlight the activity patterns that tended to be ambiguous because of measurement noise.
For this reason, the linear SVM model was used in this paper. Because
${{\bf{p}}_k} = {[{p_{1,k}},{p_{2,k}},...,{p_{N,k}}]^T}$ and ${p_{i,k}} \in \{ 0,1\} $ , the solution of ${\bf{W}}_{\bf{k}}^{\bf{T}}$ was a binary optimal problem.
The final optimal problem \cite{cortes1995support} was as below.
\begin{equation}
\begin{array}{l}
minmize:{\rm{ J(}}{{\bf{W}}_{\bf{k}}}{\rm{) = }}\frac{1}{2}{\left\| {{{\bf{W}}_{\bf{k}}}} \right\|^2}{\rm{ + }}C\sum\limits_{i = 1}^N {\xi _i^{(k)}} \\
\\
subject~{\rm{ }}to:{\rm{ }}y_i^{(k)}({\bf{W}}_{\bf{k}}^{\bf{T}}\Theta (x_i^{(k)}) + {w_0}){\rm{ }} \ge {\rm{1}}\\
\\
~~~~~~~~~~~~~~~~{\rm{                       }}\xi _i^{(k)} \ge 0,i = 1,...,N{\rm{    }}
\end{array}\
\end{equation}
Here, $\Theta $ is a linear kernel function, and $C$ is the free parameter that affects the trade-off between complexity and the number of nonseparable samples.
${\xi_i ^{(k)}}$ denotes the slack variables. Notably, for the linear SVM model, we set the self-adaptive distance parameter in the CorrNet framework as follows:
\begin{equation}\label{equ:ECS2}
{\varepsilon _{{d_j}}} = \min _{E({\bf{po}}{{\bf{s}}_j},{\bf{po}}{{\bf{s}}_m}){\rm{ = 1}}}^{m \ne j}\{ d(j,m)\}.
\end{equation}

\subsubsection{Spike neural network}

Because the learned connections of SNNs represent dynamic spatiotemporal relationships derived from fMRI signals, SNNs are more biomimetic than ANNs in both information transmission and structure.
Therefore, in this paper, a dynamic evolving SNN was also used as another pattern representation model. The SNN was built in a brain template by mapping the correlation bin ${\bf{bi}}{{\bf{n}}_k}$ into the input template of the SNN. Then, the input data were transformed into spike trains, which revealed the spatiotemporal patterns in the correlation pair $({{\bf{p}}_k},{\bf{bi}}{{\bf{n}}_k})$. 
 The strength value vector ${\bf{x}}_i^{(k)}$ for $i=1,...,N, k=1,...,D_2$ in ${\bf{bi}}{{\bf{n}}_k}$ constitutes a spike train for the $i^{th}$ trial.
 Specifically, all spike trains derived from the encoding module included $D_2$ patch states, which were injected into $D_2$ Tempotron neurons in the leaky integrate-and-fire (LIF) model \cite{brette2005adaptive}. 
 Each spike was transmitted by the synapse between the input neuron and output neuron, and each synapse had its own weight ${{\bf{w}}^{(k)}}$ equivalent to synaptic efficacy.
 The spiking neuron model can be presented as follows:
\begin{equation}
{\tau _m}\frac{{\partial v}}{{\partial t}} =  - v(t) + MI(t),t = 1,...,N,
\end{equation}
where $\tau _m$ presents the constant membrane time (we chose $\tau _m= 4\times\tau _s = 20ms$), $I_t$ is the input of a leaky integrator at $t$ time, $v(t)$ denotes the membrane potential, and $M$ is the membrane resistance. Here, the spiking threshold $v_{thre}$ is introduced.
When $v(t) \ge {v_{thre}}$ , the neuron will fire the spike, and the membrane potential will immediately reset to the reset potential $v_{rest}=0.6$.
Then, it maintains $v(t)=v_{rest}$ for a short period of time, which refers to the absolute refractory period of the biologic neuron \cite{kendrick1979testosterone}.
The subthreshold membrane potential was determined by the weighted sum of postsynaptic potentials (PSPs) from all incoming spikes:
\begin{equation}
{V^{(k)}}(t) = \sum\limits_i {w_{_i}^{(k)}\sum\limits_{{t_i}} {{K}(t - {t_i}) + {V_{rest}}} },
\end{equation}
where ${w_{_i}^{(k)}}$ and $t_i$ are, respectively, the synaptic efficacy and fire time of the $i^{th}$ afferent neuron, and $K$ is a normalized PSP kernel.
Because ${{\bf{p}}_k} = {({p_{1,k}},{p_{2,k}},...,{p_{N,k}})^T}$ and ${p_{i,k}} \in \{ 0,1\} $, two modes corresponding to the two states (0 and 1) were defined: the bright mode, and the dark mode. The learning procedure of a neuron was conducted by changing its synaptic efficacy ${w_{_i}^{(k)}}$ when error occurred.
There were two error modes: the bright error means that a neuron failed to fire when it should fire; otherwise, the dark error occurs.
In the training process, based on the analysis of different errors, the learning rule could be demonstrated as below.
\setlength{\arraycolsep}{0.5 pt}
\begin{equation}
\Delta {w_i} = \left\{ \begin{array}{ll}
lr\sum\limits_{{t_i} < {t_{\max}}} {K({t_{\max }} - {t_i})} &{,\rm{bright~error}}\\
\\
-lr\sum\limits_{{t_i} < {t_{\max}}} {K({t_{\max }} - {t_i})} &{,\rm{dark~error}}\\
\\
0&{,\rm{otherwise}}
\end{array} \right.
\end{equation}
Here, $t_{max}$ represents the time when the voltage reaches the maximum value, and $lr=0.005$ is the learning rate.
Considering that the SNN requires plenty of spikes to enable the encoding modes, we extended the ${\bf{bi}}{{\bf{n}}_k}$ 
by increasing the self-adaptive distance parameter $\varepsilon _{{d_j}}$ in the CorrNet framework:
\begin{equation}
{\varepsilon _{{d_j}}} = \frac{1}{{{D_1}}}\sum\limits_{l = 1}^{{D_1}} {d({{\bf{v}}_j},{{\bf{v}}_l})}.
\end{equation}

\section{Experiments}

\subsection{Data preparation}

Three male subjects participated in the fMRI study. They had an average age of $23$, and normal or corrected-to-normal visual acuity.
Informed written consent was obtained from all subjects.
Each subject participated in two scan sessions. The ﬁrst session included $352$ patterns with random shapes ,and the second session had $80$ patterns with regular shapes.
%
%
The random shapes of patterns were mainly used to train the decoding model, while the other group was used to test the performance of the model.
%
%
Each stimulus pattern consisted of $12\times12$ pixel patches ($1.13^\circ  \times 1.13^\circ $ each). There were two types of patches: a flickering checkerboard (spatial frequency: $1.78$ cycles/$^ \circ $; temporal frequency: $6$ Hz), and a neutral gray area. Each type of patch was randomly used with equal probability. Each pattern formed by different types of patches was presented on a neutral gray background. There was a fixation spot in the center of each stimulus, and we instructed subjects to fixate on it. There were $20$ kinds of stimulus patterns in regular shapes, including digital numbers and geometric shapes.
%
%

A rear-projection display device was used to present the stimuli outside the scanner. The resolution of the display device was $800\times600$ at $60$Hz. Subjects watched the stimuli presented by the device via a mirror placed above the subject and attached to the receptive field (RF) coil. The visual angle was $13.5^\circ$. A laptop controlled stimuli presentation using the E-prime software.
Each subject underwent two scan sessions. Random shape stimuli were displayed in the first session. There were $11$ runs in the session. Each run began with a $28$s null stimulus, followed by presenting $18$ stimulus blocks, and ending with a $12$s rest period. In each stimulus block, a $6$s stimulus period, followed by a $6$s rest period, was presented. A $30$–$60$s rest period was inserted between two runs. A total of $198$ different stimulus patterns were used in this session. Regular shape stimuli were displayed in the second session. There were ten runs, including five geometric-shape runs and five alphabet runs. The block design was the same as in the first scan session. Each specific shape stimulus was repeated six times during the whole scan session.
%
%
The color of the fixation spot in the center of each stimulus changed randomly during stimulus presentation. Each subject was asked to press a button when the color changed.

A 3.0-Tesla GE MR Scanner was used to collect functional MRI data at the First Affiliated Hospital of Xi'an Jiaotong University.
%
%
A T1-Weighted, magnetization-prepared rapid-acquisition gradient-echo (MP-RAGE) sequence (TR: $2250$ ms; TE: $2.98$ ms; Tl: $900$ ms; Flip angle: $9^\circ$; FOV: $256\times256$ mm; Voxel size: $1.0\times1.0\times1.0$ mm), was firstly used to acquire high-resolution structure images of the same slices in echo-planar imaging (EPI) sequence. Then, a T$2*$-weighted gradient-echo EPI sequence (TR: $4000$ ms; TE: $30$ ms; Flip angle: $80^\circ$; FOV: $192\times192$ mm; Voxel size: $1.875\times1.875\times3$ mm; Slice gap: $0$ mm; Number of slices: $48$) was used to collect functional images covering the whole brain.
The first three volumes of each run were discarded in order to avoid the noise caused by the MRI scanner's instability. SPM$12$ was used to preprocess the MRI data.
First, slice-timing correction was implemented to correct the differences in slice acquisition times. Second, slice-timing corrected data underwent head motion correction. Then the structure data of the first scan session were coregistered to all data in this session. Next, the structure data in the second scan session was coregistered to those in the first scan session. Finally, all data in the second scan session were coregistered to the structure data in this session.
%
%
%
%

\subsection{Performance evaluation and analysis}

The $352$ stimulus–response pairs with random shapes were used as the training set, and the other $80$ with regular shapes were used as the test set. In the experiments, to demonstrate the reliability and robustness of the CorrNet framework combined with the linear SVM (Our SVM) and the dynamic evolving SNN (Our SNN), we compared our method to baseline methods, that is, the linear SVM without the CorrNet process (Pure SVM) \cite{de2008combining} and the SNN without the CorrNet process (Pure SNN) \cite{kasabov2017mapping}. Pure SVM and Pure SNN both used all $3037$ voxels as input for pixel-wise reconstruction. The fixed-basis model \cite{miyawaki2009visual} and BCCA \cite{fujiwara2013modular} were also used for comparison. In addition, we combined our CorrNet framework with the fixed-basis model (Our Multiscale) by replacing the weights of voxels optimized by sparse logistic regression \cite{yamashita2009sparse} with the coefficients of voxels in the probabilistic correlation graph produced by our CorrNet framework when the self-adaptive distance parameter was defined by Eq. \ref{equ:ECS2}. All experiments were implemented in MATLAB R$2016$b, and used the LIBSVM Toolbox \cite{chang2011libsvm}.

\begin{figure}[htbp]
  \begin{center}
  \includegraphics[width=0.48\textwidth]{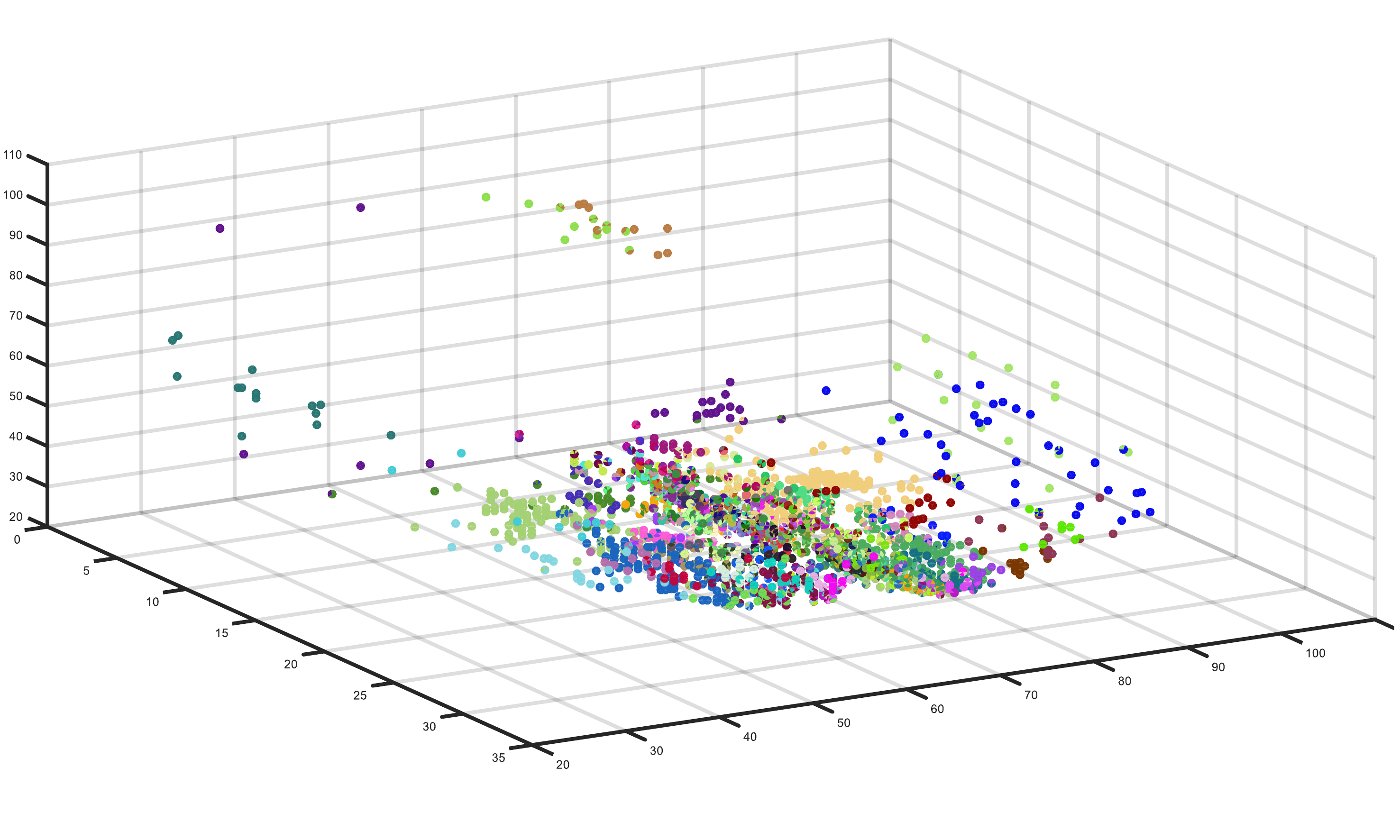}\\
  \caption{Topological distribution of all correlation bins. The voxels in the same bin are filled with the same color, and different bins are expressed in different colors.}\label{fig:tp}
  \end{center}
\end{figure}

\begin{figure}[htbp]
  \begin{center}
  \includegraphics[width=0.48\textwidth]{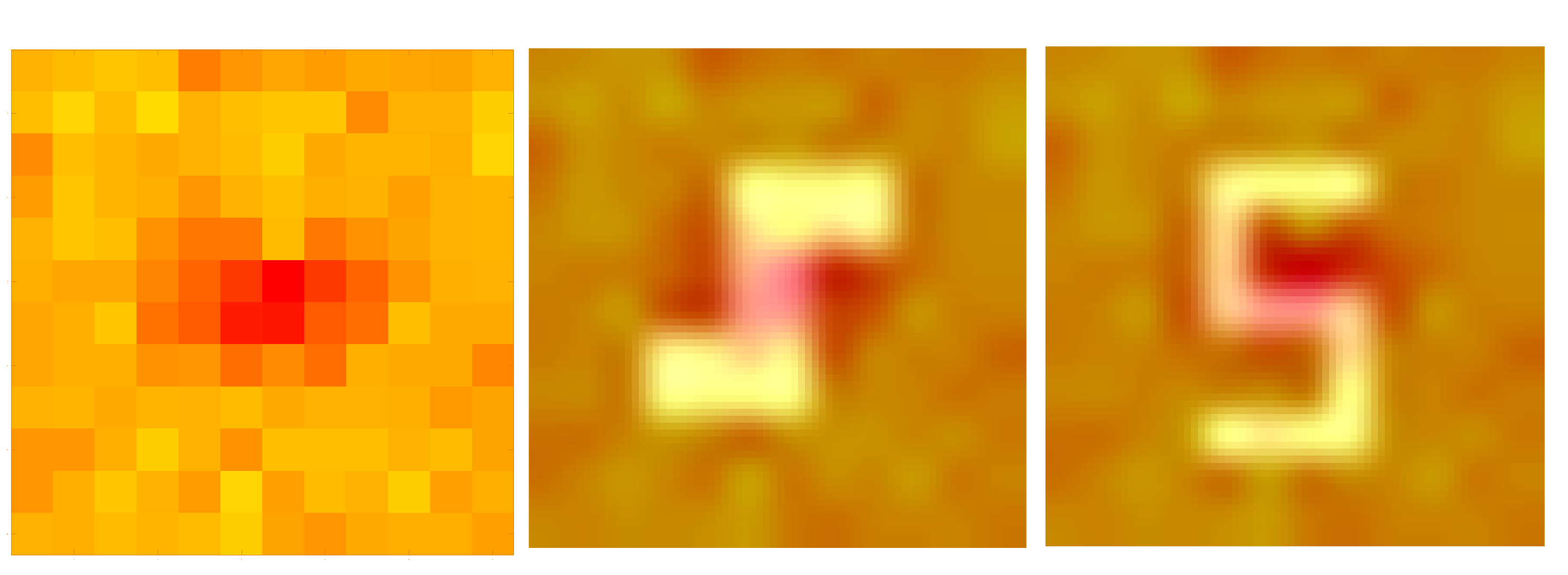}\\
  \caption{Quantitative distribution of voxels in the image coordinate frame. The first picture shows the original quantitative distribution, and the more red the pixel patch, the more voxels the corresponding correlation bin contains. The second and third picture shows the perceptual images with a quantitative distribution overlay.}\label{fig:hh}
  \end{center}
\end{figure}

Fig. \ref{fig:tp} shows the topological distribution of the $144$ correlation bins, when the self-adaptive distance parameter was defined by Eq. \ref{equ:ECS2} in the CorrNet framework.
The average number of voxels in the bins was $30$, and the utilization of all $3037$ voxels was $58\%$. Obviously, the CorrNet framework could improve the efficiency of the reconstruction method by condensing data.
Besides, a heat map was used to reveal the quantitative distribution of voxels by mapping the number of voxels in ${\bf{bi}}{{\bf{n}}_k}$ into the coordinates of ${\bf{p}}_k$ in the image coordinate frame, as shown in Fig. \ref{fig:hh}. The middle region contains the most voxels, which conforms to the attention mechanism in human perception. Therefore, our CorrNet framework could reveal the attention mechanism by combining the topological correlation with the strength correlation.

\begin{figure*}[ht]
  \begin{center}
  \includegraphics[width=0.96\textwidth]{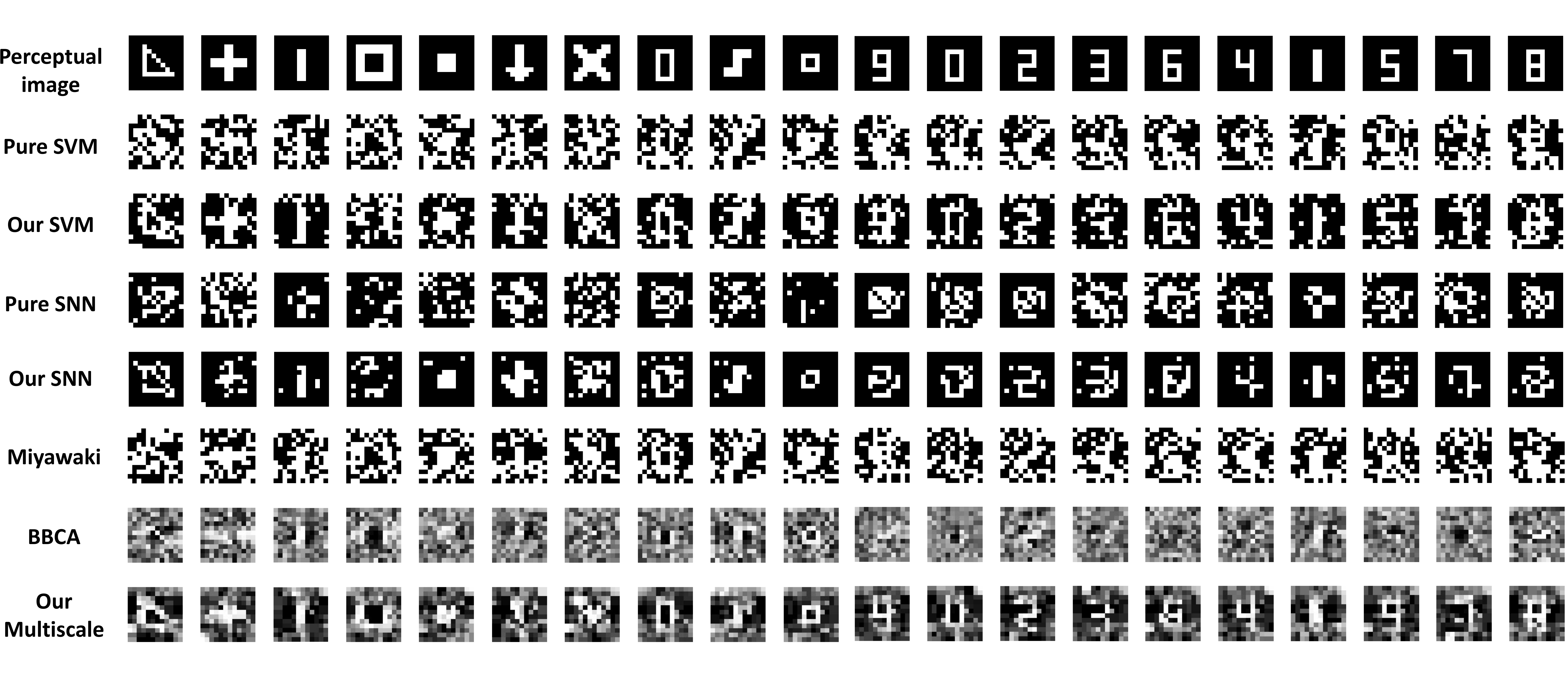}\\
  \caption{Image reconstructions of geometric shapes and digits taken from the test set in the experiments. The second to fifth rows are the binary reconstruction images, and the latter rows are the gray reconstruction images}\label{fig:reoimage}
  \end{center}
\end{figure*}

The test set includes geometric shape images and digit images with $144$ binary pixels. Notably, when compared with baseline methods, the reconstruction results of our method were binary images because the pattern representation was processed as a binary classification problem. Conversely, the results of Miyawaki, BBCA, and Our Multiscale were gray value because the pattern representation was processed as a regression problem.
%
%
%
%
The reconstruction results are shown in 
Fig. \ref{fig:reoimage}, where the first row denotes the original perceptual images, and below rows are the reconstruction images obtained from all methods in our experiments.
Obviously, the reconstruction images obtained by our methods (Our SVM, Our SNN, and Our Multiscale) were much more recognizable than those of baseline methods Miyawaki and BBCA. Specially, Our SVM and Our SNN (the third and fifth rows) achieved significant improvement over baselines (the second and fourth rows); in particularly, Our SNN achieved nearly $100\%$ accuracy in the margins of the image coordinate frame. Therefore, it can be said that the CorrNet framework indeed contributed to the visually significant improvement. As for the gray images, Our Multiscale (the last row) successfully reconstructed the relatively legible geometric shapes and digits while Miyawaki and BBCA nearly failed to separate the meaningful shapes from the background. This result strongly verifies the reliability and robustness of the proposed CorrNet framework. Besides, in the middle area, our methods all performed much better than baseline methods in reconstructing visually recognizable figures, which demonstrated that the attention mechanism could be revealed by our methods (see Figure \ref{fig:hh}).
%
%

 \begin{table*}[htbp] 
\begin{center}
\caption{ Average Accuracy of all Methods in Our Experiments.}
\label{tb:traintestav}
\begin{tabular}{c | c | c | c | c | c | c | c }
\hline
{\bf Dataset} & {\bf Pure SVM} & {\bf Our SVM} & {\bf Pure SNN} & {\bf Our SNN} & {\bf Miyawaki} & {\bf BBCA} & {\bf Our Multiscale}\\
\hline
Training set & \textbf{98.87$\pm$1.13}  & 78.71$\pm$1.04 & 84.84$\pm$2.54 & 98.02$\pm$0.23 & 95.02$\pm$0.67 & 82.06$\pm$1.30 & 84.94$\pm$0.13\\
Geometry  & 60.01$\pm$3.30 & 75.60$\pm$3.52 & 73.36$\pm$1.27 & \textbf{84.55$\pm$0.97} & 60.36$\pm$3.21 & 73.75$\pm$0.27 & 84.10$\pm$0.57\\
Digit & 57.37$\pm$3.96   & 78.09$\pm$2.03   & 78.52$\pm$1.67   &  85.61$\pm$1.62 & 56.17$\pm$3.46  & 73.18$\pm$0.96 & \textbf{85.63$\pm$0.52}\\
Test set& 58.68$\pm$3.54   & 76.85$\pm$3.13   & 75.06$\pm$1.41   & \textbf{85.11$\pm$1.22} & 58.92$\pm$3.44  & 73.46$\pm$0.57 & 84.86$\pm$0.79\\
\hline
\end{tabular}
\end{center}
\end{table*}

\begin{figure*}[!t]
\centering
\subfloat[Our SVM]{\includegraphics[width=2.3in]{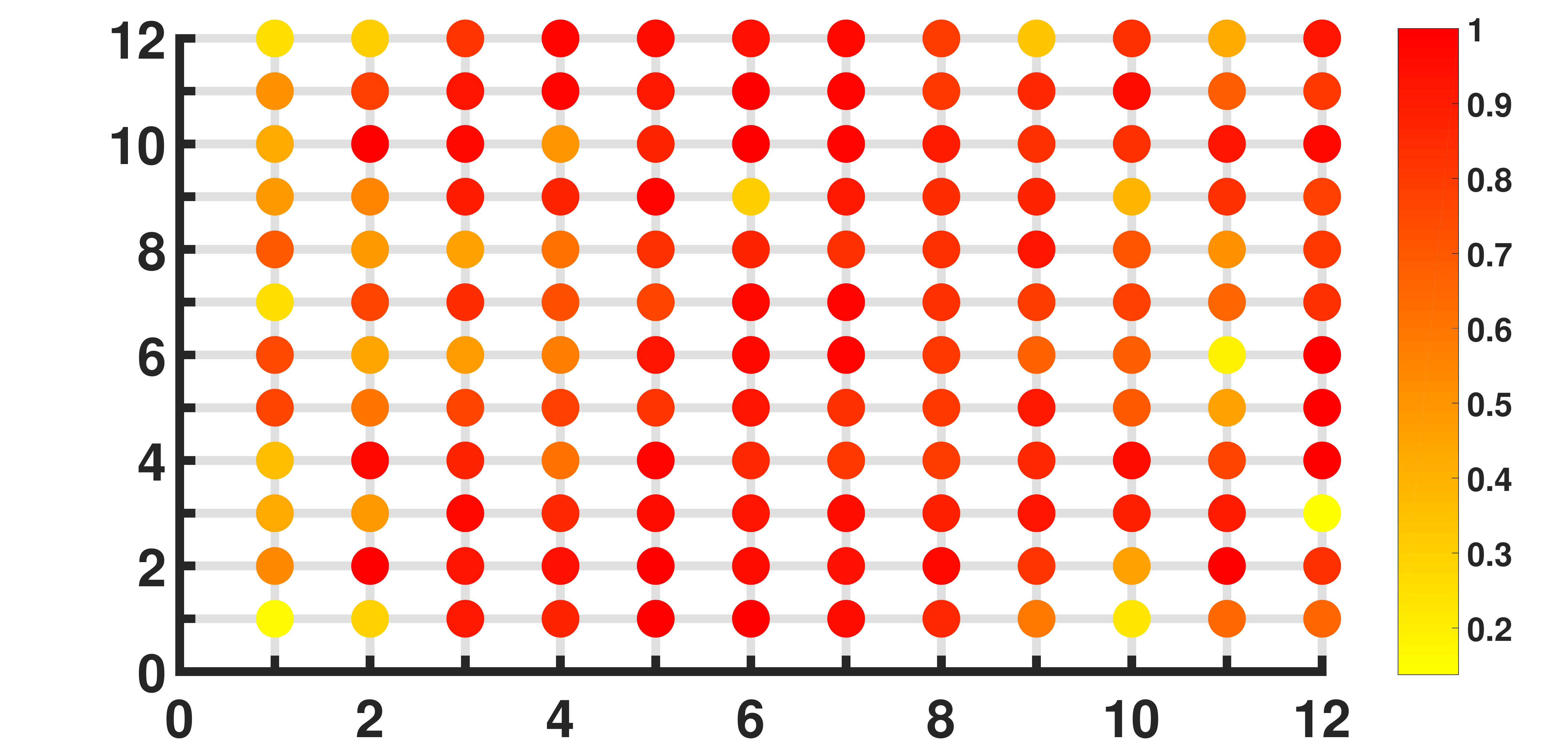}%
\label{Our SVM}}
\hfil
\subfloat[Our SNN]{\includegraphics[width=2.3in]{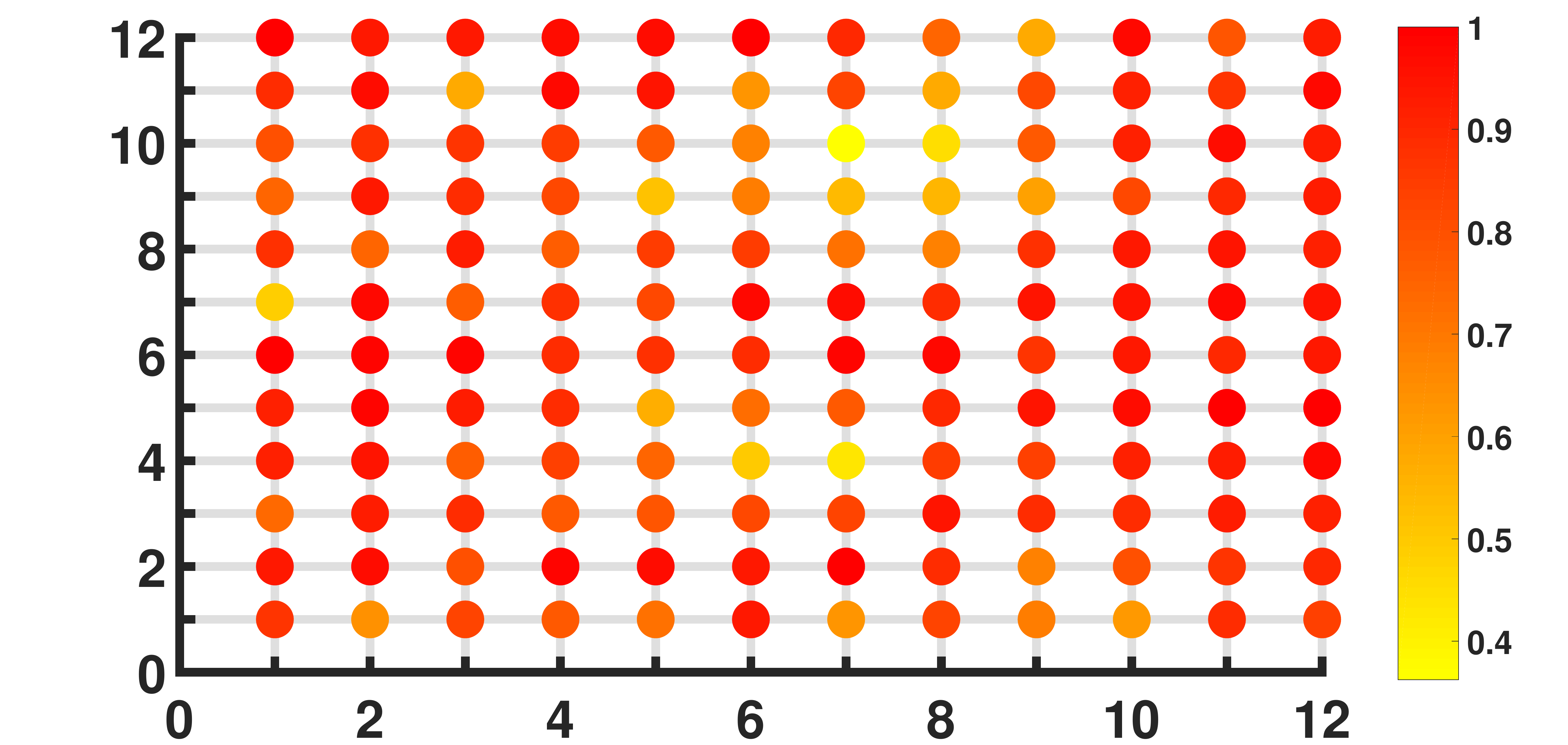}%
\label{Our SNN}}
\hfil
\subfloat[Our Multiscale]{\includegraphics[width=2.3in]{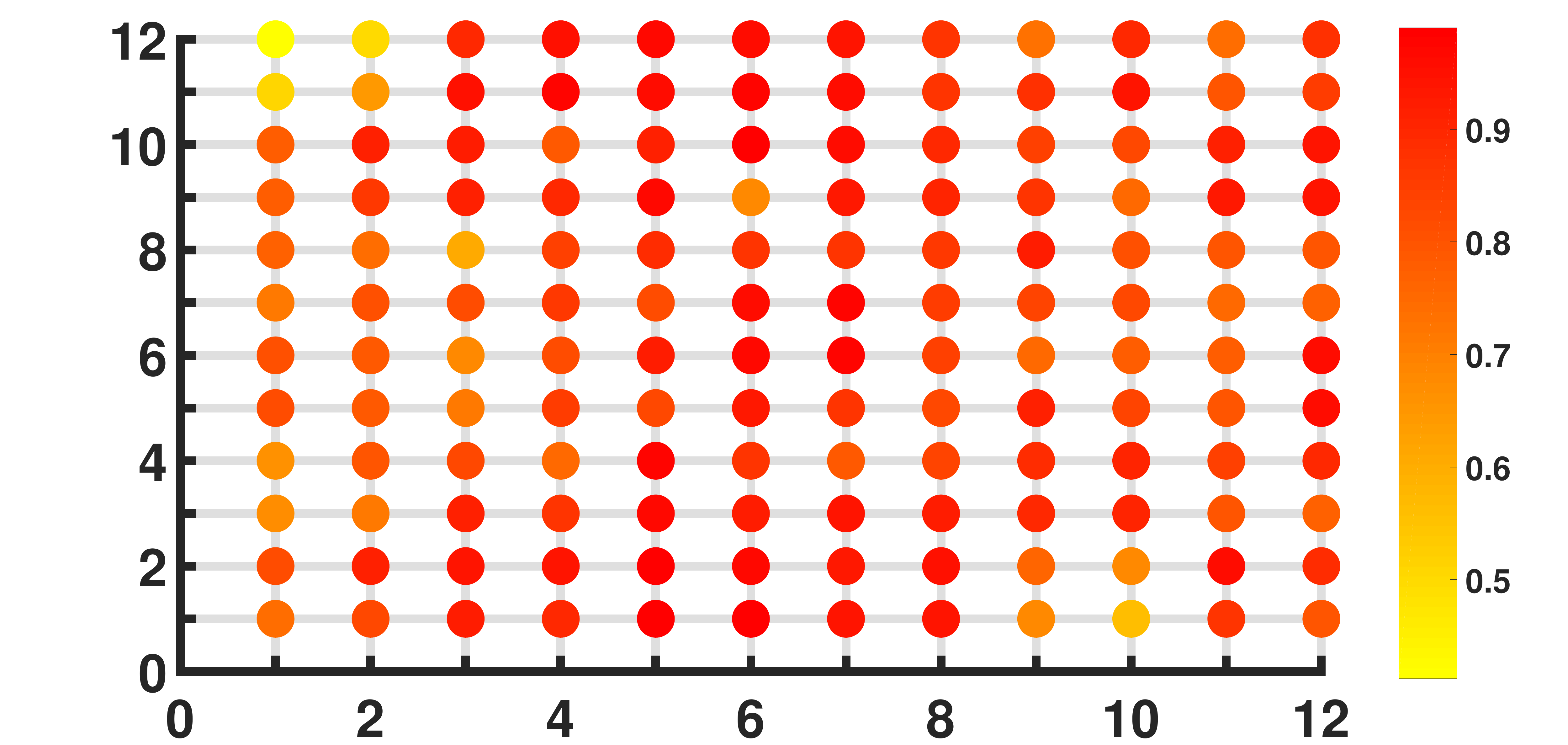}%
\label{Our Multiscale}}
\caption{The accuracy distribution of each pixel. (a)$\sim$(c) show the distribution of our SVM, our SNN, and Our Multiscale, respectively. The more red the point, the higher the accuracy of the corresponding pixel.}
\label{fig:distribution}
\end{figure*}

Referring to past reconstruction studies \cite{miyawaki2009visual,fujiwara2013modular,schoenmakers2013linear}, the reconstruction accuracy was computed pixel-by-pixel under Euclidean Distance measurements. The results on the training set, geometry, digit, and the test set are shown in Table \ref{tb:traintestav}. Both Our SVM and Our SNN achieved at least a $10\%$ improvement over the baselines (Pure SVM and Pure SNN), indicating the effectiveness of the CorrNet framework. The proposed methods could also avoid the over-fitting problem existing in the Pure SVM. Notably, for pattern representation, the SNN achieved higher average accuracy than the linear SVM, implying the prospect of SNNs in brain decoding.
%
%
In addition, comparison with Miyawaki and BBCA confirmed the superiority of our methods and CorrNet framework. The $85.11\%$ and $84.86\%$ accuracy achieved by Our SNN and Our Multiscale surpasses the state-of-the-art methods (which have around $80\%$ accuracy) in reconstructing perceptual images from fMRI signals.
%
%

Furthermore, in Fig. \ref{fig:distribution}, where the coordinate frame corresponds to the image coordinate frame, the accuracy distribution of a single reconstruction pixel on the test set is shown. The more red points means higher accuracy. Observably, if ignoring some bad points, which can be ascribed to noise, the accuracy of the pixels in the middle of the image coordinate frame is relatively high. Again, the attention mechanism was revealed. Specifically, more than $80\% $ of pixels reached $0.9$ accuracy by Our SNN, and more than $65\%$ of pixels reached $0.75$ accuracy by Our SVM and Our Multiscale. Moreover, the proportion of pixels with lower than $0.5$ accuracy was reduced to less than $15\%$ by our methods. In particular, Our SNN achieved comparably comparative accuracy in both the middle and the margin of the image coordinate frame, thus illustrating the topological robustness inside the symmetrical structure of the SNN.
%
%
\begin{figure}[htbp]
  \begin{center}
  \includegraphics[width=0.46\textwidth]{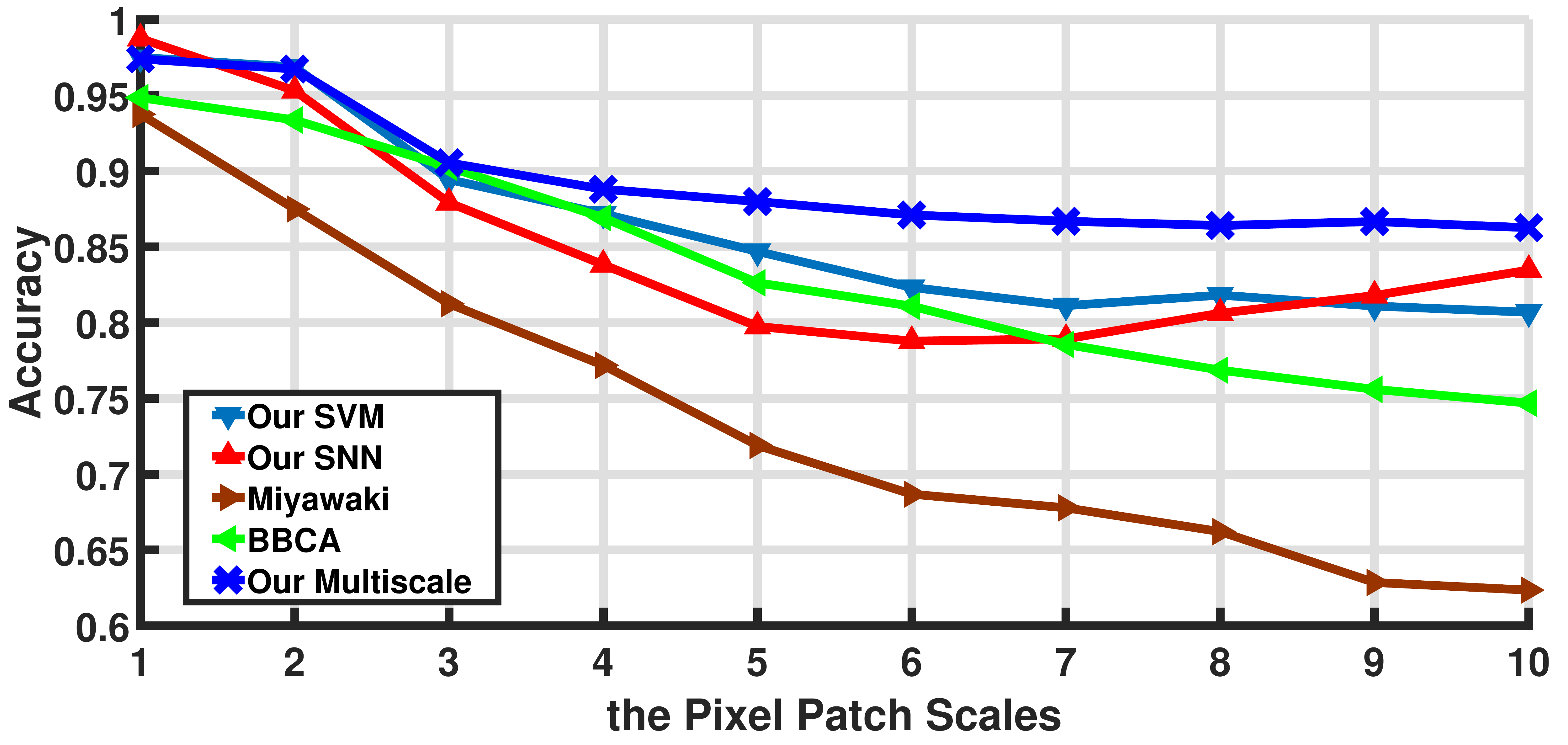}\\
  \caption{Reconstruction accuracy of pixel patches with diverse scales in our experiments. The horizontal ordinate denotes the scale values of pixel patches, and the vertical ordinate presents the accuracy.}\label{fig:curve}
  \end{center}
\end{figure}

As Fig. \ref{fig:curve} shows, we computed the pixel-patch accuracy, which denotes the average reconstruction accuracy of the pixel patches with diverse scales ( $1\times1$,..., $10\times10$ ). A pixel patch with a certain scale presents a certain region extracted from the middle to the margin in the image coordinate frame.
The accuracy curves of our methods, Miyawaki, and BBCA, were all in decline. 
However, the curves of Our SNN all had an overall uptrend, going down before going up. Visibly, considering the obviously distinct gap in accuracy between Miyawaki and Our Multiscale, the proposed CorrNet framework led to a distinct improvement on every patch scale in the fixed base method.
Our Multiscale had obvious superiority over all methods, which implies that multiscale reconstruction has great prospects in brain decoding with compound models.
%
%
%
%
%
%
Additionally, our methods tended to provide more smooth curves. This can be owed to the CorrNet framework derived from brain structure and function, which may simulate the robustness in signal process of the human brain.
%
%

\section{Conclusion}

In this paper, we proposed a CorrNet framework for brain decoding in terms of two primary problems now; this framework could be flexibly combined with diverse pattern representation models.
As far as we know, topological correlation has been fused with the strength correlation to enrich the correlations for the first time.
Then, we represented brain activity patterns by combining the linear SVM and the dynamic evolving SNN separately, with the correlation pairs derived from the CorrNet framework.
The experimental results demonstrated that our method could achieve a significant improvement in perceptual image reconstruction over baseline methods. The CorrNet framework was reliable and robust.
In addition, the attention mechanism was revealed by the new method.
Notably, for the first time, we discussed the three types of connections (functional connectivity, structural connectivity, and effective connectivity) in attention tasks based on fMRI signals.

Our work contributes to revealing the working mechanism of the human brain, simulating the effective connectivity of the early visual cortex, and establishing mapping from perceptual tasks to voxels. In engineering, this study promotes the development of efficient brain–machine interface.
Thus, future work can be focused on hybrid models fusing the CorrNet framework with diverse pattern representation models. Besides, more experiments on other fMRI datasets are required to further improve and generalize the CorrNet framework for practical applications.

\section*{Acknowledgments}

This research was partially funded by the National Natural Science Foundation of China (No. 61773312 and L1522023), and the 973 Program (No. 2015CB351703).

\ifCLASSOPTIONcaptionsoff
  \newpage
\fi

\bibliographystyle{IEEEtran}
\bibliography{bare_conf}

\begin{thebibliography}{10}
\providecommand{\url}[1]{#1}
\csname url@samestyle\endcsname
\providecommand{\newblock}{\relax}
\providecommand{\bibinfo}[2]{#2}
\providecommand{\BIBentrySTDinterwordspacing}{\spaceskip=0pt\relax}
\providecommand{\BIBentryALTinterwordstretchfactor}{4}
\providecommand{\BIBentryALTinterwordspacing}{\spaceskip=\fontdimen2\font plus
\BIBentryALTinterwordstretchfactor\fontdimen3\font minus
  \fontdimen4\font\relax}
\providecommand{\BIBforeignlanguage}[2]{{%
\expandafter\ifx\csname l@#1\endcsname\relax
\typeout{** WARNING: IEEEtran.bst: No hyphenation pattern has been}%
\typeout{** loaded for the language `#1'. Using the pattern for}%
\typeout{** the default language instead.}%
\else
\language=\csname l@#1\endcsname
\fi
#2}}
\providecommand{\BIBdecl}{\relax}
\BIBdecl

\bibitem{zheng2017hybrid}
N.-n. Zheng, Z.-y. Liu, P.-j. Ren, Y.-q. Ma, S.-t. Chen, S.-y. Yu, J.-r. Xue,
  B.-d. Chen, and F.-y. Wang, ``Hybrid-augmented intelligence: collaboration
  and cognition,'' \emph{Frontiers of Information Technology \& Electronic
  Engineering}, vol.~18, no.~2, pp. 153--179, 2017.

\bibitem{haxby2001distributed}
J.~V. Haxby, M.~I. Gobbini, M.~L. Furey, A.~Ishai, J.~L. Schouten, and
  P.~Pietrini, ``Distributed and overlapping representations of faces and
  objects in ventral temporal cortex,'' \emph{Science}, vol. 293, no. 5539, pp.
  2425--2430, 2001.

\bibitem{kamitani2005decoding}
Y.~Kamitani and F.~Tong, ``Decoding the visual and subjective contents of the
  human brain,'' \emph{Nature neuroscience}, vol.~8, no.~5, pp. 679--685, 2005.

\bibitem{hausfeld2014multiclass}
L.~Hausfeld, G.~Valente, and E.~Formisano, ``Multiclass fmri data decoding and
  visualization using supervised self-organizing maps,'' \emph{NeuroImage},
  vol.~96, pp. 54--66, 2014.

\bibitem{naselaris2015voxel}
T.~Naselaris, C.~A. Olman, D.~E. Stansbury, K.~Ugurbil, and J.~L. Gallant, ``A
  voxel-wise encoding model for early visual areas decodes mental images of
  remembered scenes,'' \emph{Neuroimage}, vol. 105, pp. 215--228, 2015.

\bibitem{norman2006beyond}
K.~A. Norman, S.~M. Polyn, G.~J. Detre, and J.~V. Haxby, ``Beyond mind-reading:
  multi-voxel pattern analysis of fmri data,'' \emph{Trends in cognitive
  sciences}, vol.~10, no.~9, pp. 424--430, 2006.

\bibitem{kay2008identifying}
K.~N. Kay, T.~Naselaris, R.~J. Prenger, and J.~L. Gallant, ``Identifying
  natural images from human brain activity,'' \emph{Nature}, vol. 452, no.
  7185, p. 352, 2008.

\bibitem{horikawa2017generic}
T.~Horikawa and Y.~Kamitani, ``Generic decoding of seen and imagined objects
  using hierarchical visual features,'' \emph{Nature communications}, vol.~8,
  2017.

\bibitem{park2013structural}
H.-J. Park and K.~Friston, ``Structural and functional brain networks: from
  connections to cognition,'' \emph{Science}, vol. 342, no. 6158, p. 1238411,
  2013.

\bibitem{miyawaki2009visual}
Y.~Miyawaki, H.~Uchida, O.~Yamashita, M.-a. Sato, Y.~Morito, H.~C. Tanabe,
  N.~Sadato, and Y.~Kamitani, ``Visual image reconstruction from human brain
  activity: A modular decoding approach,'' in \emph{Journal of Physics:
  Conference Series}, vol. 197, no.~1.\hskip 1em plus 0.5em minus 0.4em\relax
  IOP Publishing, 2009, p. 012021.

\bibitem{fujiwara2013modular}
Y.~Fujiwara, Y.~Miyawaki, and Y.~Kamitani, ``Modular encoding and decoding
  models derived from bayesian canonical correlation analysis,'' \emph{Neural
  computation}, vol.~25, no.~4, pp. 979--1005, 2013.

\bibitem{kingma2013auto}
D.~P. Kingma and M.~Welling, ``Auto-encoding variational bayes,'' \emph{arXiv
  preprint arXiv:1312.6114}, 2013.

\bibitem{wang2015deep}
W.~Wang, R.~Arora, K.~Livescu, and J.~Bilmes, ``On deep multi-view
  representation learning,'' in \emph{Proceedings of the 32nd International
  Conference on Machine Learning (ICML-15)}, 2015, pp. 1083--1092.

\bibitem{woolrich2004fully}
M.~W. Woolrich, M.~Jenkinson, J.~M. Brady, and S.~M. Smith, ``Fully bayesian
  spatio-temporal modeling of fmri data,'' \emph{IEEE transactions on medical
  imaging}, vol.~23, no.~2, pp. 213--231, 2004.

\bibitem{kuang2014discrimination}
D.~Kuang, X.~Guo, X.~An, Y.~Zhao, and L.~He, ``Discrimination of adhd based on
  fmri data with deep belief network,'' in \emph{International Conference on
  Intelligent Computing}.\hskip 1em plus 0.5em minus 0.4em\relax Springer,
  2014, pp. 225--232.

\bibitem{zeeman1965topology}
E.~C. Zeeman, ``Topology of the brain,'' 1965.

\bibitem{laconte2005support}
S.~LaConte, S.~Strother, V.~Cherkassky, J.~Anderson, and X.~Hu, ``Support
  vector machines for temporal classification of block design fmri data,''
  \emph{NeuroImage}, vol.~26, no.~2, pp. 317--329, 2005.

\bibitem{maass1996lower}
W.~Maass, ``Lower bounds for the computational power of networks of spiking
  neurons,'' \emph{Neural computation}, vol.~8, no.~1, pp. 1--40, 1996.

\bibitem{kasabov2014neucube}
N.~K. Kasabov, ``Neucube: A spiking neural network architecture for mapping,
  learning and understanding of spatio-temporal brain data,'' \emph{Neural
  Networks}, vol.~52, pp. 62--76, 2014.

\bibitem{kasabov2017mapping}
N.~K. Kasabov, M.~G. Doborjeh, and Z.~G. Doborjeh, ``Mapping, learning,
  visualization, classification, and understanding of fmri data in the neucube
  evolving spatiotemporal data machine of spiking neural networks,'' \emph{IEEE
  transactions on neural networks and learning systems}, vol.~28, no.~4, pp.
  887--899, 2017.

\bibitem{hossein2016reconstruction}
G.-A. Hossein-Zadeh \emph{et~al.}, ``Reconstruction of digit images from human
  brain fmri activity through connectivity informed bayesian networks,''
  \emph{Journal of neuroscience methods}, vol. 257, pp. 159--167, 2016.

\bibitem{engel1997retinotopic}
S.~A. Engel, G.~H. Glover, and B.~A. Wandell, ``Retinotopic organization in
  human visual cortex and the spatial precision of functional mri.''
  \emph{Cerebral cortex (New York, NY: 1991)}, vol.~7, no.~2, pp. 181--192,
  1997.

\bibitem{shmuel2007spatio}
A.~Shmuel, E.~Yacoub, D.~Chaimow, N.~K. Logothetis, and K.~Ugurbil,
  ``Spatio-temporal point-spread function of fmri signal in human gray matter
  at 7 tesla,'' \emph{Neuroimage}, vol.~35, no.~2, pp. 539--552, 2007.

\bibitem{yamashita2008sparse}
O.~Yamashita, M.-a. Sato, T.~Yoshioka, F.~Tong, and Y.~Kamitani, ``Sparse
  estimation automatically selects voxels relevant for the decoding of fmri
  activity patterns,'' \emph{NeuroImage}, vol.~42, no.~4, pp. 1414--1429, 2008.

\bibitem{yamashita2009sparse}
O.~Yamashita, ``Sparse logistic regression toolbox,'' 2009.

\bibitem{flandin2007bayesian}
G.~Flandin and W.~D. Penny, ``Bayesian fmri data analysis with sparse spatial
  basis function priors,'' \emph{NeuroImage}, vol.~34, no.~3, pp. 1108--1125,
  2007.

\bibitem{penny2011statistical}
W.~D. Penny, K.~J. Friston, J.~T. Ashburner, S.~J. Kiebel, and T.~E. Nichols,
  \emph{Statistical parametric mapping: the analysis of functional brain
  images}.\hskip 1em plus 0.5em minus 0.4em\relax Academic press, 2011.

\bibitem{beckmann2003general}
C.~F. Beckmann, M.~Jenkinson, and S.~M. Smith, ``General multilevel linear
  modeling for group analysis in fmri,'' \emph{Neuroimage}, vol.~20, no.~2, pp.
  1052--1063, 2003.

\bibitem{schoenmakers2013linear}
S.~Schoenmakers, M.~Barth, T.~Heskes, and M.~van Gerven, ``Linear
  reconstruction of perceived images from human brain activity,''
  \emph{NeuroImage}, vol.~83, pp. 951--961, 2013.

\bibitem{de2008combining}
F.~De~Martino, G.~Valente, N.~Staeren, J.~Ashburner, R.~Goebel, and
  E.~Formisano, ``Combining multivariate voxel selection and support vector
  machines for mapping and classification of fmri spatial patterns,''
  \emph{Neuroimage}, vol.~43, no.~1, pp. 44--58, 2008.

\bibitem{sporns2013network}
O.~Sporns, ``Network attributes for segregation and integration in the human
  brain,'' \emph{Current opinion in neurobiology}, vol.~23, no.~2, pp.
  162--171, 2013.

\bibitem{chew1989constrained}
L.~P. Chew, ``Constrained delaunay triangulations,'' \emph{Algorithmica},
  vol.~4, no. 1-4, pp. 97--108, 1989.

\bibitem{cortes1995support}
C.~Cortes and V.~Vapnik, ``Support vector machine,'' \emph{Machine learning},
  vol.~20, no.~3, pp. 273--297, 1995.

\bibitem{brette2005adaptive}
R.~Brette and W.~Gerstner, ``Adaptive exponential integrate-and-fire model as
  an effective description of neuronal activity,'' \emph{Journal of
  neurophysiology}, vol.~94, no.~5, pp. 3637--3642, 2005.

\bibitem{kendrick1979testosterone}
K.~M. Kendrick and R.~F. Drewett, ``Testosterone reduces refractory period of
  stria terminalis neurons in the rat brain,'' \emph{Science}, vol. 204, no.
  4395, pp. 877--879, 1979.

\bibitem{chang2011libsvm}
C.-C. Chang and C.-J. Lin, ``Libsvm: a library for support vector machines,''
  \emph{ACM transactions on intelligent systems and technology (TIST)}, vol.~2,
  no.~3, p.~27, 2011.

\end{thebibliography}

\end{document}